\documentstyle[colacl]{article}

\newcommand{\disco}{{\sc Disco}}
\newcommand{\shdg}{{\sc shdg}}
\newcommand{\tdl}{{\sc TDL}}
\newcommand{\udine}{{\sc UDiNe}}
\newcommand{\morphix}{{\sc Morphix-3}}
\newcommand{\sereal}{{\sc SeReal}}

\newcommand{\attrib}[1]{\mbox{\tt #1\ }}
\newcommand{\emptynode}[0]{\mbox{$[\,]$}}

\newcommand{\attribute}[1]{{\tt #1}}
\newcommand{\avm}[1]{\mbox{\begin{math}
                           \setlength{\arraycolsep}{1mm}
                           \hspace*{-0.35em} 
                           \left[ 
                           \begin{array}{@{}l@{~}l@{}}          
                             \\[-0.16in] #1 \\[-0.16in]           
                           \end{array}
                           \right] \hspace*{-0.05em}
                           \end{math}}}
\newcommand{\attval}[2]{\mbox{\attribute{#1}} & {\tt #2}  \\ }
\newcommand{\emptylist}[0]{\mbox{$\left\langle ~\right\rangle$}}

\newcommand{\ind}[1]{\fbox{{\scriptsize #1}}\, }
\newcommand{\nelist}[1]{\mbox{$\left\langle #1 \right\rangle$}}

\newlength{\itemlen} 

{\begin{list}{}{
  \setlength{\labelwidth}{\itemlen}
  \setlength{\labelsep}{0em}
  \setlength{\leftmargin}{\itemlen}
  \setlength{\parsep}{1ex}
  \setlength{\listparindent}{0em}}}%
{\end{list}}

\newlength{\descilen} 
\author{Stephan Busemann\\
DFKI GmbH \\ 
Stuhlsatzenhausweg 3, D-66123 Saarbr\"{u}cken, Germany\\
e-mail: {\tt busemann@dfki.de}
}
\title{Interfacing Constraint-Based Grammars and Generation Algorithms}
\date{}

\begin{document}
\bibliographystyle{acl}
\pagestyle{empty}
\maketitle
\begin{abstract}
Constraint-based grammars  can,
in principle, serve as the major linguistic knowledge source for both
parsing and generation.
Surface generation starts from input semantics representations that may vary
across grammars. For many declarative grammars, the concept of
derivation implicitly built in is that of parsing. They may thus
not be interpretable by a generation algorithm. 
We show that linguistically plausible semantic analyses can cause
severe problems for semantic-head-driven approaches for generation 
(\shdg). We use \sereal, a variant of \shdg\ and the
\disco\ grammar of German as our source of
examples. We propose a new, general approach that explicitly accounts for the
interface between the grammar and the generation algorithm by adding a
control-oriented layer to the linguistic knowledge base that
reorganizes the semantics in a way suitable for generation. 

\end{abstract}
\section{Introduction}
The relation between declaratively represented grammars and control
structures that can process them is often described along the following
lines: The declarative representation of grammars allows a grammar
writer to describe
the well-formed sentences of a language in an non-directional way.
Such grammars can be used for both parsing and generation and are
called reversible. On the other hand, algorithms
for processing grammars are free from language-specific stipulations and
can operate on different grammars within the same formalism.

Unfortunately, this picture is too superficial and needs some discussion and
refinement. The declarative representation of grammars is a necessary
prerequisite for ensuring reversibility, but it is far from
sufficient. 

This paper reports on a case study, in which a declaratively
represented grammar developed by linguists and used
for parsing was employed for generation. The generation algorithm used
is a variant of Semantic-Head-Driven Generation
(\shdg) \cite{Shi:Noo:Moo:90}. It interprets a large constraint-based
grammar of German developed for the \disco\ system  (Dialogue System for 
Cooperating agents) \cite{Usz:Bac:Bus:94}. \shdg\
is one of the most widespread algorithms for 
sentence realization with constraint-based grammars. It is largely 
theory-independent and has been used for Head-Driven Phrase Structure 
Grammars (HPSG), Definite Clause Grammars, and Categorial Unification 
Grammars. Since its publication, \shdg\ had to compete with other algorithms 
(e.g.\ \cite{Rus:War:Car:90}, \cite{Strzalkowski:94b},
\cite{Mar:Str:92}) which led to numerous ways of improving the basic
procedure. 

A major question remained unsolved (and it is
unsolved for other algorithms as well), namely that of the  algorithm's 
requirements on the properties of the grammar used. In previous work,
Shieber imposed a condition on ``semantic monotonicity'' that holds for
a grammar if  for every phrase  the semantic structure of each immediate
subphrase subsumes some portion of the semantic structure of the entire
phrase \cite[p.~617]{Shieber:88}. Semantic monotonicity is very strict and
could be relaxed in \shdg: It was shown that semantically
non-monotonic grammars can be processed by \shdg. 
It is a yet open question whether all semantically monotonic grammars
can be processed by \shdg\ and what the class of \shdg-processable
grammars is.

Using the linguistically well motivated semantics of  \disco\
as a sample input language to \shdg, we show that there are
semantically monotonic grammars 
that cannot be processed directly by \shdg. The difficulties encountered
are of a general kind, and a general approach for solving them is
presented that explicitly accounts for the 
interface between the grammar and the generation algorithm by adding a
modular, control-oriented layer to the linguistic knowledge base that
represents a reorganization of the semantics in a way suitable for
generation. Moreover, we present the specific solution for the grammar
in hand.   

The kind of problem investigated in this paper relates to the
fundamental question of how to organize a modular system consisting of
linguistic knowledge (a grammar) and control knowledge (parser or
generator). It turns out that declarative grammars contain hidden 
assumptions about processing issues that need to be made explicit.

\section{SHDG and the Grammar Interface}

Without loosing generality we assume that the grammar has a
semantics layer and that a generator input expression is an element of
the semantic representation language encoded by the grammar. The
generator is guided by its input. Thus we
refer to the semantics layer as the {\em essential feature}. 

We now briefly review some essential points of \shdg\ \cite{Shi:Noo:Moo:90}. 
The algorithm is centered around the notion of a {\em pivot\/} node, which 
provides an essential feature specification from which it first generates all 
descendants in a top-down manner, and then tries to connect the newly 
generated subtree to a higher node (or the root node) in bottom-up fashion.
Both generating descendants and connecting to higher nodes involves the 
application of grammar rules. Correspondingly, 
rules are subdivided into two classes:
{\em chain rules\/} are used for bottom-up connection 
while {\em non-chain rules\/} are
applied for top-down expansion. Chain rules differ from non-chain rules in 
that their left-hand side essential feature is identical to the
essential feature
of one of their right-hand side elements. This element
is called the ``semantic head'' of the chain rule.
Lexical entries are non-chain rules in a trivial way since they have no
categorial right-hand side elements.

The only specific assumption \shdg\ makes about a grammar is that 
chain rules and their semantic
heads can be identified. However, the property of being a chain rule 
(or non-chain rule) is often assigned by the grammar writer on purely 
linguistic grounds although it determines the processing strategy: 
If the set of chain rules happens
to be empty, \shdg\ operates strictly top-down. 
If the set of non-chain rules consists of lexicon entries only, 
\shdg\ behaves like a bottom-up generator. 
Having the linguist unconsciously influence the processing strategy of
\shdg\ can lead to uninterpretable grammars, as we will show below.

We now introduce some basic assumptions about grammars.
A grammar induces a context-free backbone and has
separate layers to represent morphological, syntactic, and semantic
properties of categories. We assume furthermore that the
generator can be told how to identify mother and daughter categories of
grammar rules. 

It has always been a matter of discussion how a surface generator
should cope with the presence or absence of essential feature
specifications. Since a closer investigation of this issue is
beyond the scope of this paper, we assume that the input structure and
the semantics of the sentence to be generated are equal, i.e.\ they
subsume each other.

A generator must {\em terminate\/} with results on all allowable
input. The under-specification
of the essential feature at execution time is a well-known phenomenon
\cite{Rus:War:Car:90}. It can show up during top-down expansion
of a grammar rule that does not share
the essential features of the daughters with parts of the mother.
Non-termination or failure to find a derivation will result.
To avoid underspecification, the following condition on
generator/grammar pairs ensures successful recursive
applicability of the generation procedure:

\begin{quote}
{\bf Essential Feature Specification Condition (EFSC):} The 
essential feature must specify exactly the constituent to be generated
at the time the generation procedure is executed on it.
\end{quote}

This  requirement captures what should be stated independently of particular
generators and grammars. For individual generator/grammar pairs, EFSC
needs to be concretized. Any such concretization must take the specific
algorithm into account. For instance,
specifications of EFSC involving \shdg\ depend on  the order in which nodes of 
a local tree are recursively expanded. \cite{Shi:Noo:Moo:90} 
quite arbitrarily assume a strict left-to-right
processing of non-semantic-head daughter nodes. EFSC is easily violated
by a daughter of a non-chain rule that influences the essential feature
of a preceding daughter.

For the purpose of the present paper, we make a simplification by
assuming that ``exact specification'' implies that the essential
feature specification is subsumed by (i.e.\ is as specific as) the
corresponding part of the input  structure. While this is a sufficient
condition on successful termination, the necessary one may indeed be
weaker. It depends on the architecture of the grammar, from which we
want to abstract away. 

\begin{figure*}[h,t]
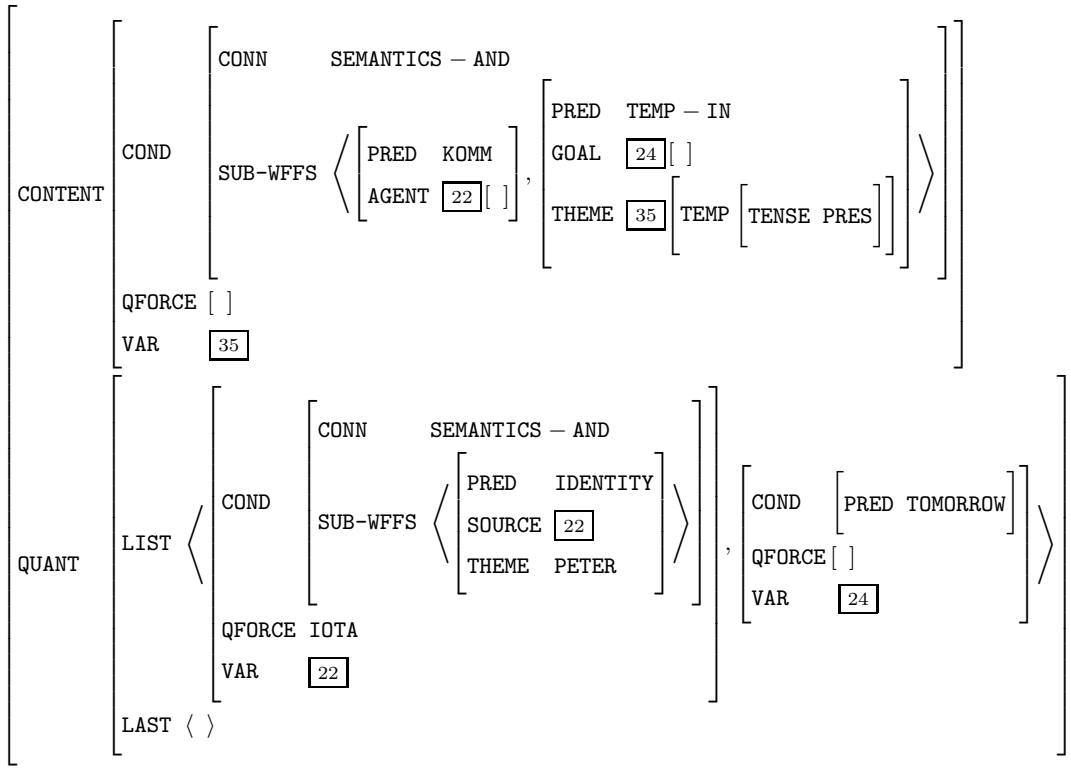

\renewcommand{\baselinestretch}{1.5}
\begin{center}
\small{
\avm{\attval{CONTENT}{\avm{\attval{COND}{\avm{\attval{CONN}{SEMANTICS-AND}
                                              \attval{SUB-WFFS}{
\nelist{\; \avm{\attval{PRED}{KOMM}
   \attval{AGENT}{\ind{22}\; \avm{}}     },\;
   \avm{\attval{PRED}{TEMP-IN}     \attval{GOAL}{\ind{24}\; \avm{ }}  
\attval{THEME}{\ind{35}\, \avm{\attval{TEMP}{\avm{\attval{TENSE}{PRES}
         }}}} } }  } }} 
                           \attval{QFORCE}{\; \avm{}}
                           \attval{VAR}{\ind{35}}
                          } }
     \attval{QUANT}{\avm{\attval{LIST}{
\nelist{\; \avm{\attval{COND}{\avm{\attval{CONN}{SEMANTICS-AND} 
      \attval{SUB-WFFS}{\nelist{\; \avm{\attval{PRED}{IDENTITY}
      \attval{SOURCE}{\ind{22}}
      \attval{THEME}{PETER} } } } }}
         \attval{QFORCE}{IOTA}
     \attval{VAR}{\ind{22}} },\; \avm{\attval{COND}{\avm{\attval{PRED}{TOMORROW}} } 
    \attval{QFORCE}{\avm{}} 
    \attval{VAR}{\ind{24}}  } } }  
    \attval{LAST}{\emptylist}     }  }
    }
}
\end{center}
\renewcommand{\baselinestretch}{1}

\caption{Semantic Feature Structure for {\em Peter
kommt morgen\/} [Peter arrives tomorrow].}
\label{sample} 
\end{figure*}

\section{The System Setup}

This section introduces the generator/grammar pair used for the present
study. After a  sketch of our variant of \shdg\ we discuss the
semantics layer of the constraint-based grammar of German to the extent
necessary to demonstrate violation of EFSC and to describe a solution.

\subsection{The \sereal\ system}

The \sereal\ 
(Sentence Realizer) is a Common Lisp  \shdg\ implementation that 
uses kernel components of the \disco\ NL understanding system
\cite{Usz:Bac:Bus:94}.

\disco\ is a linguistic core engine capable of analyzing NL sentences
as quasi-logical form representations that can subsequently be 
submitted to further semantic analysis.  
The \disco\  grammar is encoded in
\tdl\ \cite{Kri:Sch:94}, a powerful type definition language 
and type inference mechanism for feature structures. The basic
processing engine is the feature constraint solver \udine, which is used to 
perform (destructive) unification during parsing and generation. 
A mapping between word forms and morpho-syntactically annotated word stems
is achieved by the  \morphix\ system \cite{Fin:Neu:88}. 
\sereal\ is integrated into the \disco\ system to the extent that it 
uses the same grammar, \udine, \tdl, and \morphix. It can 
be fed with the parser's semantics output and thus serve as a
useful grammar development tool.

A special mechanism had to be developed for efficient lexicon
access. The \shdg\ 
algorithm simply assumes all lexicon entries to be available as
non-chain rules. This is, however, not advisable for large lexicons. Rather,
only the relevant entries should be accessed. Therefore, \sereal\
indexes the lexicon according to semantic information. 
Consider, for instance, the semantic representation in
Figure~\ref{sample}.\footnote{This is a  simplified
version of a semantic representation taken from a parse with the
\disco\ grammar. For presentation purposes we adopt the familiar 
matrix notation for feature structures. $<$ and $>$ are print macros for lists
that expand into the common feature structure notation for lists (cf.\
\cite[page 29]{Shieber:86a}). Although \tdl\ defines {\em typed\/} feature
structures, we omit type information here as it is not relevant.}
Lexical indices usually are semantic predicates
denoted by the {\tt PRED} feature, e.g. {\tt KOMM} is the index
for the main verb ({\em arrive\/}). Exceptions include 
determiners, which are indexed according to the value of {\tt QFORCE} and
proper names, which are indexed according to the value of {\tt
THEME}. A priority system on indices ({\tt THEME $>$ QFORCE $>$ PRED}) 
reduces the number of accessible
indices. This way an index points to very few lexicon
entries.\footnote{This depends on how many
lexemes carry the same index. Usually we have one to three, in rare
cases up to fifteen, entries per index.} Indices are retrieved as values
of some path in the essential feature specification.
Insertion of an entry into a derivation requires its essential feature
to subsume the input structure to avoid introducing spurious semantic
specifications. 

Clearly both indices and path
descriptions are grammar dependent and form a part of the interface 
between \sereal\ and the \disco\ grammar.
In Figure~\ref{sample}, the following indices are used to
access lexicon entries: {\tt KOMM, PETER, TEMP-IN}.

\begin{figure*}[h,t]
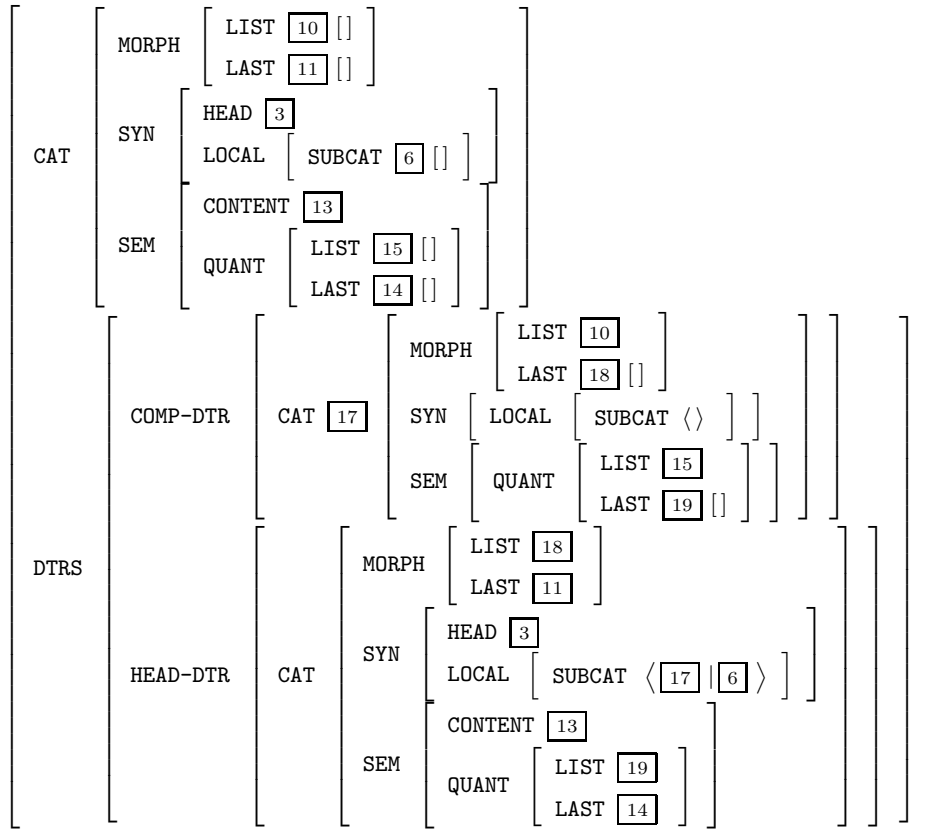

\renewcommand{\baselinestretch}{1.5}
{\small
\begin{displaymath}
\left[\begin{array}{l}
\attrib{CAT}
\left[\begin{array}{l}

\attrib{MORPH}
\left[\begin{array}{l}
\attrib{LIST}\ind{10}\,\emptynode\\
\attrib{LAST}\ind{11}\,\emptynode
\end{array}\right]\\

\attrib{SYN}
\left[\begin{array}{l}
\attrib{HEAD}\ind{3}\\
\attrib{LOCAL}
\left[\begin{array}{l}
\attrib{SUBCAT}\ind{6}\,\emptynode
\end{array}\right]
\end{array}\right]\\

\attrib{SEM}
\left[\begin{array}{l}
\attrib{CONTENT}\ind{13}\\

\attrib{QUANT}
\left[\begin{array}{l}
\attrib{LIST}\ind{15}\,\emptynode\\
\attrib{LAST}\ind{14}\,\emptynode
\end{array}\right]
\end{array}\right]
\end{array}\right]\\

\attrib{DTRS}
\left[\begin{array}{l}
\attrib{COMP-DTR}
\left[\begin{array}{l}
\attrib{CAT}\ind{17}\,
\left[\begin{array}{l}
\attrib{MORPH}
\left[\begin{array}{l}
\attrib{LIST}\ind{10}\,\\
\attrib{LAST}\ind{18}\,\emptynode
\end{array}\right]\\
\attrib{SYN}
\left[\begin{array}{l}
\attrib{LOCAL}
\left[\begin{array}{l}
\attrib{SUBCAT}\langle\,\rangle\
\end{array}\right]
\end{array}\right]\\

\attrib{SEM}
\left[\begin{array}{l}
\attrib{QUANT}
\left[\begin{array}{l}
\attrib{LIST}\ind{15}\,\\
\attrib{LAST}\ind{19}\,\emptynode
\end{array}\right]
\end{array}\right]
\end{array}\right]
\end{array}\right]\\

\attrib{HEAD-DTR}
\left[\begin{array}{l}
\attrib{CAT}
\left[\begin{array}{l}

\attrib{MORPH}
\left[\begin{array}{l}
\attrib{LIST}\ind{18}\,\\
\attrib{LAST}\ind{11}\,
\end{array}\right]\\

\attrib{SYN}
\left[\begin{array}{l}
\attrib{HEAD}\ind{3}\,\\
\attrib{LOCAL}
\left[\begin{array}{l}
\attrib{SUBCAT}\left\langle\,\ind{17}\,|\,\ind{6}\,\right\rangle
\end{array}\right]
\end{array}\right]\\
\attrib{SEM}
\left[\begin{array}{l}
\attrib{CONTENT}\ind{13}\,\\
\attrib{QUANT}
\left[\begin{array}{l}
\attrib{LIST}\ind{19}\,\\
\attrib{LAST}\ind{14}\,
\end{array}\right]
\end{array}\right]
\end{array}\right]
\end{array}\right]
\end{array}\right]
\end{array}\right]
\end{displaymath}
}  

\renewcommand{\baselinestretch}{1}
\caption{A Head-Complement Rule (simplified for expository purposes).}
\label{hfr1} 
\end{figure*}

The algorithm has been criticized for not terminating on left-recursive
rules \cite{Strzalkowski:94b}. Under the assumption of semantic monotonicity,
the determination of a pivot can be conditioned
by a check for semantic content. If the semantics is
``empty'' (i.e., it corresponds to the top feature structure),
processing fails and alternative possibilities have to be explored. Since
left recursion occurs only in top-down direction, we are dealing with
non-chain rules,
which ensures that the semantics of a right-hand side element differs from 
that of the left-hand side. Semantic monotonicity ensures that it is
``smaller'' in some sense, thus guaranteeing termination. 

\mbox{}\cite{Mar:Str:92} criticized the possible failure of top-down 
expansion due to the strict 
left-to-right processing of the list of right-hand side elements.  
Since the instantiation of the semantics
of some right-hand elements can depend on the previous successful expansion 
of others, a strict order that does not consider such relations is
inadequate. In \sereal, the left-most right-hand side element of a rule
is expanded first that has a non-empty semantics instantiated.

\subsection{The \disco\ semantics layer}

The \disco\ grammar is a semantically monotonic lexicalized, HPSG-style
grammar of German  with about 20 rules,
13 of them binary. The remaining ones are unary (lexical) rules that serve to
introduce syntactic features for lexemes in particular environments.
For instance, verb lexemes can be made finite or infinite, adjectives
can be made attributive or predicative. The binary rules account for
complement and adjunct realization.  

The development of the \disco\ grammar was, as many others, based 
on purely linguistic motivations. 
Although a declarative representation is used, the concept of
derivation implicitly built in is that of (bottom-up) 
parsing. Again, this is common. 
The parsing view of the grammar developer influences the goals
that a semantic representation should fulfill. The \disco\
semantics layer should
\begin{itemize}
\item represent a linguistically well motivated (surface) propositional
semantics of NL sentences,
\item  provide the interface to subsequent non-compositional, 
extra-grammatical semantic
interpretation (e.g.\ anaphora resolution, scope disambiguation), and
\item represent the essential feature for grammar-based sentence realization.
\end{itemize}

The semantics layer corresponds to quasi-logical forms
\cite{Alshawi:92b} that are defined through the grammar and represented
with help of feature structures \cite{Nerbonne:92a}.
The relevance of the surface ordering of complements and adjuncts during
later semantic processing 
made it necessary to encode ordering information at the semantics
layer. This is reflected by the {\tt QUANT} feature, which
contains a list of the semantics of the complements and adjuncts
in the order they occur at the surface. The relations between them are
expressed by the {\tt CONTENT} feature with help of the {\tt VAR}
feature. 

Consider as an example the semantics structure in Figure~\ref{sample}.
{\tt QUANT} has two elements, the first one representing the 
proper name and the second one the temporal adverb {\em tomorrow}. 
{\tt CONTENT} 
represents a {\tt COND}ition on the meaning consisting of a conjunction
of sub-formulae. The first formula represents a one-place predicate
{\tt KOMM}, the argument of which points, via {\tt VAR}, 
into the first element of the {\tt
QUANT} list. The second sub-formula represents a two-place predicate {\tt
TEMP-IN}. Its first argument points into the second element of {\tt
QUANT}, and its second argument relates to the whole {\tt CONTENT}
feature. Thus the predicate is to be interpreted as a temporal
sentential modifier.

Semantic information mainly originates 
from lexical entries. A few general principles
of feature distribution are represented with the grammar
rules. Figure~\ref{hfr1} shows a head-complement rule with the
complement being the first element of the head's subcategorization
list. The complement is preceding the head (not shown).
{\tt CONTENT} is shared between the mother ({\tt CAT}) and the head daughter.
In a rule's left-hand side constituent, {\tt QUANT} denotes the concatenation
of the {\tt QUANT} values of the sequence of right-hand side elements.

List concatenation
is encoded using difference lists. Thus it is not necessary to use
functional feature values such as {\tt append}.
The difference list type built into
in \tdl\ denotes a list L by defining a list L1 under the feature {\tt
LIST} and another list L2 under the feature {\tt LAST} such that L2 is
a tail of L1 and the concatenation of L and L2 yields L1. This can
easily be achieved by choosing appropriate coreferences. 

In  the case of bottom-up processing, this mechanism is used like a stack:
at the mother node, the {\tt QUANT} feature of
the complement semantics has been pushed onto the list of elements collected
so far (at the head daughter).

\section{A Violation of EFSC}
\label{difflist}
Investigation of the grammar rules shows that 
there are no binary chain rules since the {\tt QUANT} feature within {\tt SEM}
differs at all nodes of a rule (cf.\ Figure~\ref{hfr1}). With the
resulting top-down strategy the 
{\tt QUANT} list at the mother node must be split into two sublists in order
to instantiate the {\tt QUANT} lists of the
daughter nodes. This is a nondeterministic problem that,
given the present implementation of difference lists, leads to
under-specification. 

Unification of some input semantics with the mother node (in
Figure~\ref{hfr1} under {\tt
CAT.SEM}\footnote{We use the period between feature names to denote
feature path descriptions.}) does not specify how the {\tt QUANT} list
should be split, i.e.\ the {\tt QUANT.LAST} feature
of the {\tt COMP-DTR} semantics, which is shared with the {\tt
QUANT.LIST} feature of the {\tt HEAD-DTR} semantics, is not affected at
all by this unification operation.
Any further expansion steps using similar rules will not specify the
semantics any further, and hence non-termination results.\footnote{It 
may be argued that the {\tt CONTENT} feature could serve as a
pivot. It is indeed shared between mother and head in most rules, which would
then be chain rule candidates. However, semantic information necessary to guide
the generation of many phrasal constituents may be represented only by 
{\tt QUANT}.}

This problem is not specific to the \disco\ grammar. Difference lists
are a common descriptive device used in many constraint-based
grammars. For instance, the same problem arises with the Minimal
Recursion Semantics, a framework for semantics within HPSG, which was
developed to simplify transfer and generation for machine translation
\cite{Cop:Fli:Mal:95,Cop:Fli:Sag:97}.

Neither is the problem specific to \sereal\ or \shdg. It is due to the
fact that there is no inverse function for list concatenation, causing
the simulating difference list mechanism to fail on splitting lists.

\section{Reorganizing Semantic Information}

Whenever a grammar/generator pair violates EFSC, 
two basic repair strategies offer themselves as remedies: 
Either the generator is modified to account for the
grammatical analysis, or the grammar is adapted to the needs of the generator.

\begin{figure}[t,h]
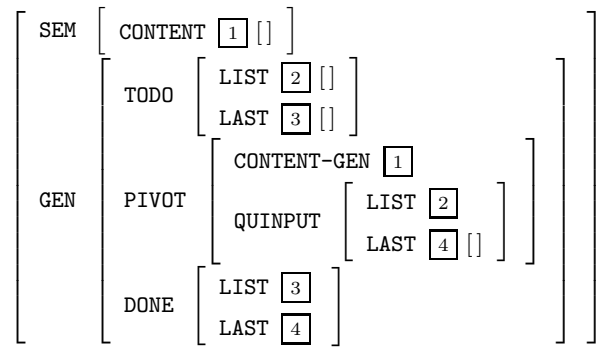

\renewcommand{\baselinestretch}{1.5}
{\small
\begin{displaymath}
\left[\begin{array}{l}
\attrib{SEM}
\left[\begin{array}{l}
\attrib{CONTENT}\ind{1}\,\emptynode 
\end{array}\right]\\
\attrib{GEN}
\left[\begin{array}{l}
\attrib{TODO}
\left[\begin{array}{l}
\attrib{LIST}\ind{2}\,\emptynode\\
\attrib{LAST}\ind{3}\,\emptynode
\end{array}\right]\\
\attrib{PIVOT}
\left[\begin{array}{l}
\attrib{CONTENT-GEN}\ind{1}\,\\
\attrib{QUINPUT}
\left[\begin{array}{l}
\attrib{LIST}\ind{2}\,\\
\attrib{LAST}\ind{4}\,\emptynode
\end{array}\right]
\end{array}\right]\\
\attrib{DONE}
\left[\begin{array}{l}
\attrib{LIST}\ind{3}\,\\
\attrib{LAST}\ind{4}\,
\end{array}\right]
\end{array}\right]
\end{array}\right]
\end{displaymath}
} 
\renewcommand{\baselinestretch}{1}
\caption{The Organization of the {\tt GEN}  Layer.}
\label{gen} 
\end{figure}

\begin{figure*}[t]
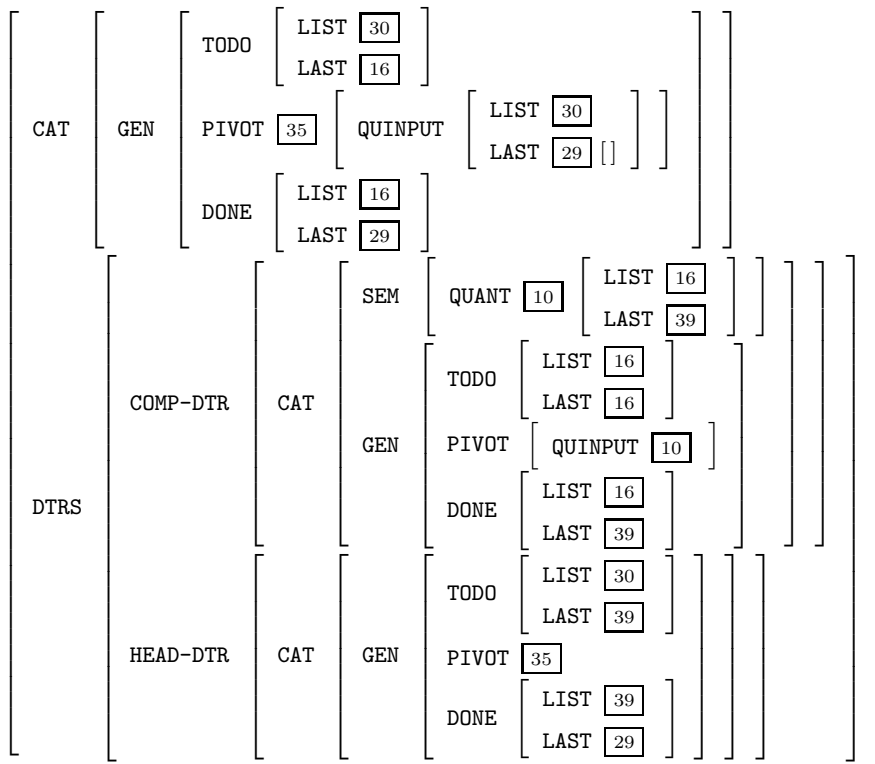

\renewcommand{\baselinestretch}{1.5}
{\small
\begin{displaymath}
\left[\begin{array}{l}
\attrib{CAT}
\left[\begin{array}{l}
\attrib{GEN}
\left[\begin{array}{l}
\attrib{TODO}
\left[\begin{array}{l}
\attrib{LIST}\ind{30}\\
\attrib{LAST}\ind{16}\,
\end{array}\right]\\

\attrib{PIVOT}\ind{35}\,
\left[\begin{array}{l}
\attrib{QUINPUT}
\left[\begin{array}{l}
\attrib{LIST}\ind{30}\,\\
\attrib{LAST}\ind{29}\,\emptynode
\end{array}\right]
\end{array}\right]\,\\

\attrib{DONE}
\left[\begin{array}{l}
\attrib{LIST}\ind{16}\,\\
\attrib{LAST}\ind{29}\,
\end{array}\right]
\end{array}\right]\\
\end{array}\right]\\

\attrib{DTRS}
\left[\begin{array}{l}
\attrib{COMP-DTR}
\left[\begin{array}{l}
\attrib{CAT}
\left[\begin{array}{l}

\attrib{SEM}\,
\left[\begin{array}{l}
\attrib{QUANT}\ind{10}\,
\left[\begin{array}{l}
\attrib{LIST}\ind{16}\,\\
\attrib{LAST}\ind{39}\,
\end{array}\right]
\end{array}\right]\\

\attrib{GEN}
\left[\begin{array}{l}

\attrib{TODO}
\left[\begin{array}{l}
\attrib{LIST}\ind{16}\,\\
\attrib{LAST}\ind{16}\,
\end{array}\right]\\


\attrib{PIVOT}
\left[\begin{array}{l}
\attrib{QUINPUT}\ind{10}
\end{array}\right]\\

\attrib{DONE}
\left[\begin{array}{l}
\attrib{LIST}\ind{16}\,\\
\attrib{LAST}\ind{39}\,
\end{array}\right]
\end{array}\right]
\end{array}\right]
\end{array}\right]\\
\attrib{HEAD-DTR}
\left[\begin{array}{l}
\attrib{CAT}
\left[\begin{array}{l}
\attrib{GEN}

\left[\begin{array}{l}

\attrib{TODO}
\left[\begin{array}{l}
\attrib{LIST}\ind{30}\,\\
\attrib{LAST}\ind{39}\,
\end{array}\right]\\

\attrib{PIVOT}\ind{35}\,\\

\attrib{DONE}
\left[\begin{array}{l}
\attrib{LIST}\ind{39}\,\\
\attrib{LAST}\ind{29}\,
\end{array}\right]
\end{array}\right]
\end{array}\right]
\end{array}\right]
\end{array}\right]
\end{array}\right]\\
\end{displaymath}
}
\renewcommand{\baselinestretch}{1}
\caption{The {\tt GEN} Feature in a Head-Complement Rule.}
\label{hfr1gen} 
\end{figure*}

Grammar writing should be guided by linguistic adequacy considerations rather
than by algorithmic issues. Linguistically plausible analyses should
not be rejected because they are not processed by the generator used.
On the other hand, designers of generation (or parsing) 
algorithms want to create generic tools that can be used for large classes of
grammars. Such algorithms, including those of the \shdg\ type, 
should not be geared towards a particular grammar. Moreover, in a large
grammar,  processing problems may occur with several phenomena, 
and solving them either way might eventually sacrifice the modularity
of the grammar and the generator. 

In conclusion, neither of the two ways is satisfactory. 
A third strategy is to design generators in such a way that they comply
with a particular grammar theory. \cite{Wil:Mat:98} describe
modifications of bottom-up generation \cite{Noord:90} to comply with the
current version of HPSG \cite{Pol:Sag:94}. Given the many variants of
linguistic theories used in implemented systems world-wide, this
strategy will probably in most cases boil down to 
adapting generators to grammars that adhere to a ``local'' version of a
theory.

%
In this contribution we present a novel approach  that complements
a single grammar by an explicit and modular interface layer that
restructures the semantic information in 
such a way that it supports bottom-up processing within \sereal.
This method improves over previous approaches in various ways:
\begin{itemize}
\item The interface is defined declaratively;
\item Reversibility properties of the grammar are preserved;
\item The modularity of the grammar and the generator are preserved.
\end{itemize}
This layer, {\tt GEN}, is assigned to every category of 
the grammar (cf.\  Figure~\ref{gen}). We make use of the
constraint-based formalism (here: \tdl) in defining {\tt GEN} and
relating it to the grammar. It is very important to notice
that its definition does not modify
the grammar, rather a new module is added to it. Since semantic
information is not constrained, but just restructured in {\tt GEN}, 
reversibility properties of the grammar are not touched. Parsing
results are completely independent from the presence of  {\tt GEN}.
Since the restructuring is achieved by using coreferences with the
parts of the semantic layer, generation uses the same kind of semantic
information as parsing. Hence, \sereal\ will deliver all sentences for
a semantic representation restructured in {\tt GEN} that yield that
semantic representation when they are parsed.

In the case at hand, we relate {\tt GEN} to the \disco\ semantics as follows.
Within {\tt GEN} we define a new essential feature, {\tt PIVOT}, that
shares the semantic content (under {\tt
CONTENT-GEN}) and contains the  {\tt QUANT} list of the  input (under {\tt
QUINPUT}). We specify explicitly the sublist of {\tt QUINPUT}
covered by the subtree represented by the category at hand using the
list {\tt DONE}, and we also note the list of remaining elements that
still need to be processed  ({\tt TODO}). This is encoded 
using difference lists. 

The binary grammar rules are extended as follows (Figure~\ref{hfr1gen}
shows
the {\tt GEN} feature added to the rule in  Figure~\ref{hfr1}). 
Mother and head daughter 
share their {\tt PIVOT} features, which yields us chain rules (and the desired
bottom-up processing strategy).
Obviously the mother's {\tt DONE} list must be
the concatenation of all daughters' {\tt DONE} lists.
Moreover, the complement daughter's
{\tt TODO} list must be empty, which is why {\tt QUINPUT} and
{\tt DONE} coincide. {\tt QUINPUT} of the complement daughter 
is shared with {\tt
SEM.QUANT}. It is completely specified
after the subtree represented by the head daughter
has been completed.

\section{Conclusion}

Interfaces between constraint-based grammars and generation systems
ultimaltely are defined in a very specialized way.  In
view of the disadvantages of current approaches dealing with EFSC
violations, we have introduced into the descriptive framework  a new, 
control-oriented layer of representation, {\tt GEN}, that reorganizes semantic
information in such a way that it does not violate EFSC for the generation
algorithm used. 

{\tt GEN} is the essential feature of a generation procedure and 
serves to define the interface between a grammar and
a generator. This way, the interface is explicitly and declaratively
defined. Grammars developed independently of a specific generator can
be adapted quickly without changing them. Different interfaces can
adapt a grammar to different generators. 

Besides its architectural advantages, this approach  
has considerable practical benefits compared to compilation methods.
It uses the same representational
means that serve for the implementation of the grammar.
If a grammar writer chooses to modify the encoding of certain linguistic
phenomena, potential clashes with the interface definitions can be
detected and removed more easily. 

The method is generally applicable in constraint-based frameworks. The
{\tt GEN} layer must be defined explicitly for every generator/grammar pair. 
Depending on whether and where EFSC is violated, {\tt GEN} may just
co-specify the semantics (the trivial case), or reconstruct the semantics in an
EFSC-compatible fashion. An instance of the latter was described above 
for the \disco\ grammar and \sereal. If a different generator is chosen for the
\disco\ grammar, neither the algorithm nor the grammar needs to be
modified. The same holds true, if \sereal\ was to interpret a different
grammar. In both cases, it
is the definition of {\tt GEN} that would have to be replaced.

The techniques presented are implemented in \tdl\ and CommonLisp
within the \sereal\ system. 

\section*{Acknowledgments}
This work has been partially supported by a grant from 
The German Federal Ministry for Research and Technology (FKZ~ITW~9402). I am
grateful to Feiyu Xu for implementing a first version of the interface,
and to Jan Alexandersson, Edmund Grimley Evans, and Harald Lochert 
for implementing parts of the \sereal\ system.

\end{document}